\documentclass[letterpaper, 10 pt, conference]{ieeeconf}
\usepackage[utf8]{inputenc}
\usepackage{standalone}
\IfStandalone{
\usepackage[top=2cm,bottom=2.5cm,left=2.5cm,right=2.5cm]{geometry}}{
}

\usepackage{times}
\usepackage{multicol}

\usepackage{tabularx}

\usepackage{amsthm,amsmath}

\usepackage{mathtools}
\usepackage{amssymb}
\usepackage{cite}
\usepackage[hidelinks]{hyperref}

\usepackage{cleveref}
\Crefname{equation}{Eq.}{Eqs.}
\Crefname{figure}{Fig.}{Figs.}
\Crefname{tabular}{Tab.}{Tabs.}
\Crefname{definition}{Def.}{Defs.}
\Crefname{section}{Sec.}{Sects.}
\Crefname{theorem}{Thm.}{Thms.}
\Crefname{lemma}{Lem.}{Lems.}
\Crefname{property}{Prop.}{Props.}
\Crefname{algorithm}{Alg.}{Algs.}

\usepackage[usenames,dvipsnames]{xcolor} %
\usepackage{graphicx}
\usepackage{booktabs}
\usepackage{threeparttable}
\usepackage{multirow}
\usepackage{tikz, pgfplots, pgfplotstable}
\usepackage{svg}
\usepackage{xspace} 

\usetikzlibrary{positioning,shapes,calc,arrows.meta}
\pgfplotsset{compat=1.15}

\usepackage{blindtext}
\usepackage[textsize=footnotesize]{todonotes}
\presetkeys{todonotes}{inline}{}
\usepackage{algorithm}
\usepackage[noend]{algpseudocode}
\usepackage{pgf}
\usepackage{pdfpages}
\usepackage{comment}

\makeatletter
\let\MYcaption\@makecaption
\makeatother
\usepackage[font=footnotesize]{subcaption}
\makeatletter
\let\@makecaption\MYcaption
\makeatother

\usepackage[acronym]{glossaries}

\usepackage{ifxetex}
\ifxetex
\usepackage[tracking=false,kerning=false,spacing=false]{microtype}
\usepackage[protrusion=true,tracking=false,kerning=false,spacing=false]{microtype}
\else
\usepackage[tracking=false,kerning=true,spacing=true]{microtype}

\fi
\let\labelindent\relax
\usepackage{enumitem}

\setlist[enumerate]{itemsep=0mm,leftmargin=4mm,topsep=0mm}
\setlist[itemize]{itemsep=0mm,leftmargin=4mm,topsep=0mm}

\renewcommand{\baselinestretch}{0.96}

\newcommand{\oneplayercolor}{rgb, 255:red, 239; green, 248; blue, 33 }
\newcommand{\twoplayercolor}{rgb, 255:red, 236; green, 120; blue, 83 }
\newcommand{\threeplayercolor}{rgb, 255:red, 155; green, 23; blue, 158 }

\newcommand{\alg}{\mathtt{ALG}}

\newcommand{\rrt}{\mathtt{RRT}}
\newcommand{\rrtstar}{\rrt^*}
\newcommand{\prm}{\mathtt{PRM}}
\newcommand{\prmstar}{\prm^*}
\newcommand{\rrg}{\mathtt{RRG}}
\newcommand{\factrrg}{\mathtt{FactRRG}}
\newcommand{\factprm}{\mathtt{FactPRM^*}}
\newcommand{\speed}{v}
\newcommand{\bitstar}{\mathtt{BIT^*}}
\newcommand{\fmtstar}{\mathtt{FMT^*}}

\newcommand{\nodeset}{X}
\newcommand{\factnearest}{\agents \text{-} \nearest}

\newcommand{\eventone}{A}

\newcommand{\drrtstar}{\mathtt{dRRT}^*}

\newacronym{nash}{NE}{Nash Equilibrium}
\newacronym{udg}{UDG}{Urban Driving Game}
\newacronym{cudg}{CUDG}{Communal Urban Driving Game}
\newacronym{gnep}{GNEP}{Generalized Nash Equilibrium}
\newacronym{ibr}{IBR}{iterated-best-response}
\newacronym{av}{AV}{Autonomous Vehicle}
\newacronym{dp}{DP}{Dynamic Programming}
\newacronym{pomdp}{POMDP}{Partially Observable Markov Decision Process}
\newacronym{ma-mpp}{MA-MMP}{Multi-Agent Motion Planning Problem}
\newacronym{mpp}{MMP}{Motion Planning Problem}
\newacronym{sba}{SBA}{Sampling-based Algorithm}
\newcommand{\init}{\mathrm{init}}
\newcommand{\rand}{\mathrm{rand}}
\newcommand{\near}{\mathrm{near}}
\newcommand{\goal}{\mathrm{goal}}

\newcommand{\freespace}{\statespace_{\mathrm{free}}}
\newcommand{\obstaclespace}{\statespace_{\mathrm{obs}}}
\newcommand{\mypath}{\sigma} %
\newcommand{\jointpath}{\mypath}
\newcommand{\pathspace}{\Sigma}
\newcommand{\problem}{\mathcal{P}}

\newcommand{\pointset}{S}
\newcommand{\disp}{D} %
\newcommand{\minN}{N}
\newcommand{\pbar}{\bar{p}}
\newcommand{\dispbar}{\bar{\disp}}
\newcommand{\gfact}{\mathsf{g}^\fact}

\newcommand{\facth}{h}
\newcommand{\factratio}{\mathsf{f}}

\newcommand{\getsample}{\mathtt{SAMPLE}}
\newcommand{\nearest}{\mathtt{NEAR}}
\newcommand{\collfree}{\mathtt{COLLISION\_FREE}}
\newcommand{\addedge}{\mathtt{ADD\_EDGE}}
\newcommand{\addnode}{\mathtt{ADD\_NODE}}

\newcommand{\factorize}{\mathtt{FACTORIZE}}
\newcommand{\statespace}{\mathcal{X}} %
\newcommand{\state}{x} %

\newcommand{\fact}{\mathsf{Fact}} %
\newcommand{\sol}{\textsc{Sol}}
\newcommand{\false}{\textsc{False}}

\newcommand{\hypergraph}{\overrightarrow{\mathcal{H}}}
\newcommand{\graph}{\mathcal{G}}

\newcommand{\agents}{\mathcal{A}} %
\newcommand{\agent}{i}
\newcommand{\agenttwo}{j}

\newcommand{\cost}{c}

\newcommand{\powerset}{\mathsf{pow}}
\newcommand{\partition}{\mathsf{Part}}
\newcommand{\prob}{\mathbb{P}} %

\newcommand{\closure}{\mathsf{cl}} %
\newcommand{\bigo}{\mathcal{O}}
\DeclarePairedDelimiter\ceil{\lceil}{\rceil}

\newcommand{\coll}{\mathsf{coll}}
\newtheorem{theorem}{Theorem}
\newtheorem{lemma}[theorem]{Lemma}

\newtheorem{definition}{Definition}
\newtheorem{proposition}{Proposition}

\newtheorem{assumption}{Assumption}

\makeatletter
\def\@opargbegintheorem#1#2#3{\trivlist
	\item[]{\bfseries #1\ #2\ (#3)} \itshape}
\makeatother
\newcommand{\reals}{\mathbb{R}}
\newcommand{\naturals}{\mathbb{N}}

\newcommand{\norm}[1]{\left\lVert#1\right\rVert}

\newcommand{\lebesgue}{\mu}

\DeclareMathOperator*{\argmin}{arg\,min}
\newcommand{\card}[1]{\left\vert#1\right\vert}
\usepackage[normalem]{ulem}

\IEEEoverridecommandlockouts                              %
\begin{document}
\title{\LARGE \bf
Factorization of Multi-Agent Sampling-Based Motion Planning%
}

\author{Alessandro Zanardi$^{1^\ast}$, Pietro Zullo$^{1^\ast}$, Andrea Censi$^{1}$, Emilio Frazzoli$^{1}$
\thanks{$^{*}$Equal contribution.}
\thanks{$^{1}$A. Zanardi, P. Zullo, A. Censi, and E. Frazzoli are with the Institute for Dynamic Systems and Control, ETH Z\"urich, Switzerland {\tt azanardi@ethz.ch}.
This work was supported by the Swiss National Science Foundation under NCCR Automation, grant agreement 51NF40\_180545.
}
}%

\maketitle
\thispagestyle{empty}

\begin{abstract}
Modern robotics often involves multiple embodied agents operating within a shared environment.
Path planning in these cases is considerably more challenging than in single-agent scenarios.
Although standard \acrfullpl{sba} can be used to search for solutions in the robots' joint space,  this approach  quickly becomes computationally intractable as the number of agents increases.
To address this issue, we integrate the concept of \emph{factorization} into sampling-based algorithms, which requires only minimal modifications to existing methods.
During the search for a solution we can decouple (i.e., \emph{factorize}) different subsets of agents into independent lower-dimensional search spaces once we certify that their future solutions will be independent of each other using a factorization heuristic. 
Consequently, we progressively construct a lean hypergraph where certain (hyper-)edges split the agents to independent subgraphs. In the best case, this approach can reduce the growth in dimensionality of the search space from exponential to linear in the number of agents. On average, fewer samples are needed to find high-quality solutions while preserving the optimality, completeness, and anytime properties of \acrshortpl{sba}.
We present a general implementation of a factorized \acrshort{sba}, derive an analytical gain in terms of sample complexity for $\prmstar$, and showcase empirical results for $\rrg$. 

\end{abstract}
\section{Introduction}

The study of \acrfullpl{ma-mpp} has gained increasing importance due to their potential to address intricate real-world problems that arise in fields such as autonomous driving and logistics.
A critical aspect of planning in shared environments lies in discovering efficient and robust coordination methods that allows for seamless navigation in the presence of other agents, just like humans do every day on the road.

As multiple agents aim to achieve individual objectives while avoiding collisions, \acrshortpl{ma-mpp} naturally emerge as the mathematical representation of this challenge. Interestingly, many successful motion planning methods, which perform well in the single-agent scenario, struggle to scale effectively in multi-agent situations~\cite{LaValle2006, StandleyCompleteAlgoCooperativePath}. A widely acknowledged limiting factors is the ability to manage search problems in high-dimensional spaces~\cite{HopCroftPSPACEhard}.  

Sampling-based algorithms have been extremely successful in solving planning problems in relatively low-dimensional search spaces~\cite{Karaman2011Sampling-basedPlanning}.
However, they are not immune to the curse of dimensionality imposed by multi-agent systems. The search space increases exponentially with the number of agents, rendering the problem  intractable for event a small number of agents. 
\begin{figure}[htb!]
\begin{subfigure}[b]{0.49\columnwidth}
  \centering
    \includegraphics[scale=1,width=\columnwidth]{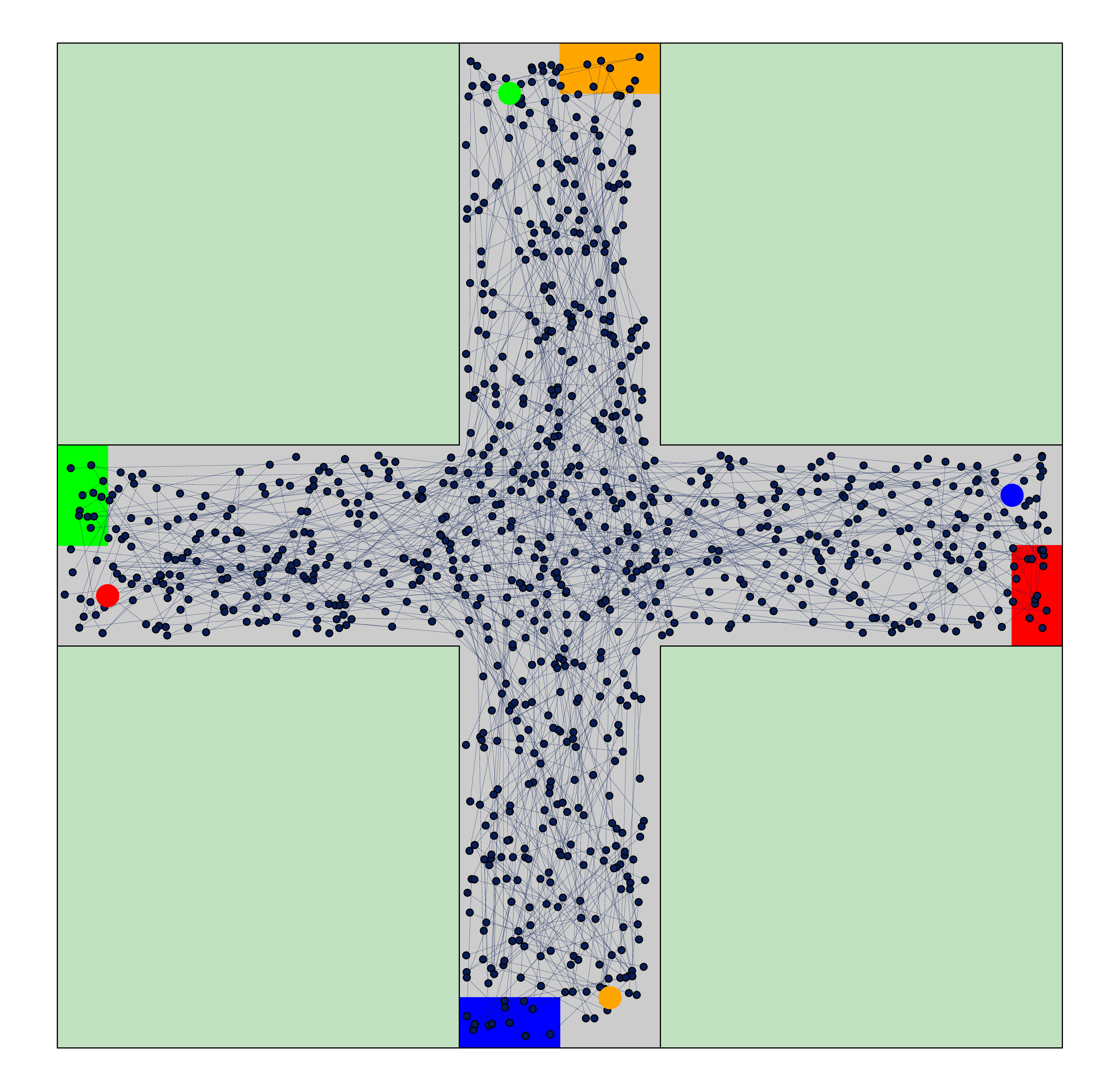}
\end{subfigure}%
\begin{subfigure}[b]{0.49\columnwidth}
  \centering
    \includegraphics[scale=1,width=\columnwidth]{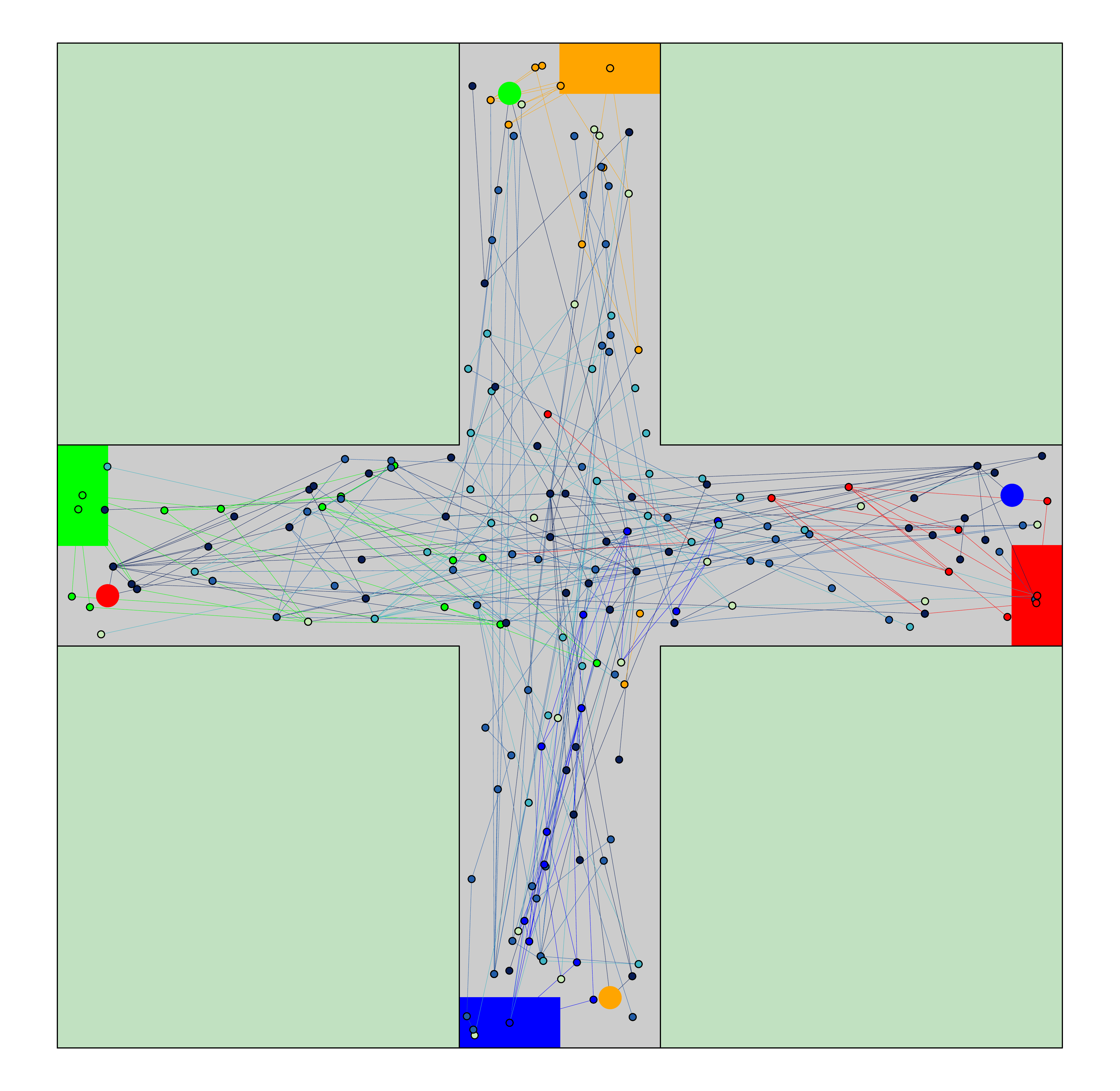}
\end{subfigure}%
\caption{A $4$-agent \acrshort{ma-mpp} example. Each agent starting from one corridor has to reach its goal region in an opposite corridor. On the left the result of from running $\rrg$ in the joint configuration space. On the right we employ the \emph{factorized} counterpart ($\factrrg$) where the players get factorized to separate lower dimensional graphs as they become independent (i.e., when we are ensured that their future solutions from that point onward are independent). One can visually appreciate that $\factrrg$ converges to a leaner graph requiring (on average) many fewer iterations to achieve an equivalent solution in cost.}
\end{figure}

Numerous attempts have been made to overcome these limitations. 
Some introduce heuristics that transform the \acrshort{ma-mpp} into a sequence of single-agent problems, for example, by imposing a specific order on the agents~\cite{multiplemovingobjectslozanoperez,MCapPrioritizedPlanning} or by developing local collision resolution methods~\cite{Silver2005CooperativePathfinding,drrtstar}. These methods often compromise optimality or completeness, particularly when agents navigate small, crowded environments~\cite{Wagner2015SubdimensionalPlanning}.
\paragraph*{Statement of Contribution}
In this work, we demonstrate how the concept of \emph{factorization}, introduced in~\cite{ZanardiFactorization} for dynamic games, can be applied to sampling-based algorithms for \acrshortpl{ma-mpp}.
Our proposed approach enables the natural decoupling of players from the joint configuration space into lower-dimensional search spaces---one for each agent at the limit---as the solution is computed. 
One main advantage of our approach is its applicability to most common off-the-shelf state-of-the-art planners. Moreover, it provides potentially significant computation gains (depending on how much the \acrshort{ma-mpp} is intrinsically factorizable) while leveraging all the properties (optimality, completeness, etc.) and tools developed for the single-agent case (e.g., extension to dynamical systems).
\subsection{Related Work}\label{sec:relatedwork}
In the pursuit of solving \acrshortpl{ma-mpp}, we identify three main approaches: \emph{centralized}, \emph{decentralized}, and \emph{hybrid}. 
\emph{Centralized} approaches treat the union of all the agents as a single system~\cite{MCapMARRT,latombecentralizedvsdecentralized,multiagentgnep}. These methods plan in the joint space and generally preserve the same completeness and optimality properties  as their corresponding single-agent counterparts. 
However, this leads to an exponential growth of the dimensionality of the search space, which quickly becomes computationally intractable~\cite{LaValle2006}. 
An example is~\cite{MCapMARRT}, which adapted~$\rrtstar$ to solve the \acrshort{ma-mpp}.

\emph{Decentralized} approaches attempt to find a solution to the \acrshort{ma-mpp} by first solving individual \emph{single-agent} problems, and subsequently composing their solutions~\cite{multiplemovingobjectslozanoperez,MCapPrioritizedPlanning,Silver2005CooperativePathfinding,drrtstar}.
Naturally, the first step ignores the coupling among the agents, and
additional steps are are needed to resolve the conflicts when compositing the individual solutions. Proposed approaches range from enforcing ``planning hierarchies'' among the agents~\cite{multiplemovingobjectslozanoperez,MCapPrioritizedPlanning} to local conflict resolution or re-planning~\cite{Silver2005CooperativePathfinding,drrtstar} after solving the single problems.
For instance,~\cite{multiplemovingobjectslozanoperez} constructs paths in sequential order, treating previous agents as moving obstacles. In~\cite{Silver2005CooperativePathfinding} paths for each agents are found individually, and if two agents collide at a certain point, the remainder of their routes is re-planned.
While computationally more efficient, these methods generally lack the same optimality and completeness guarantees of their \emph{centralized} counterparts.

Lastly, \emph{hybrid} approaches aim to combine the best of both \emph{centralized} and \emph{decentralized} approaches.
Similarly to the method we hereby present, this class of algorithms often leverages data structures and algorithms that primarily with the individual problems of the single agents but can switch to a joint representation when conflicts arise. An example is given by~\cite{drrtstar}, which introduces $\drrtstar$, an asymptotically optimal sampling-based algorithm that scales to high number of agents without requiring an explicit representation of the joint motion graph.
Often, a key insight is the ability to identify smaller independent (sub)problems that can be solved independently.
For instance,~\cite{StandleyOAID} initially computes optimal paths for each agent individually. If the individual paths are found to conflict, the conflicting agents are grouped together, and the solution is computed again, treating grouped agents as single entities. 
Similarly, \cite{Wagner2015SubdimensionalPlanning} considers collision between individual policies as an indicator of independence (or better, dependence), but includes this concept in the path construction phase, resulting in a framework that can be adapted to incremental planners such as $\rrt$ and $\prm$. 

\section{Preliminaries}\label{sec:preliminaries}

\subsection{Multi-Agent Motion Planning Problem}
We begin by recalling the \acrfull{ma-mpp} definition akin to~\cite[Chapter~7.2]{LaValle2006}.
We consider a finite set of agents $\agents$ with cardinality $\card{\agents}$.
Each agent has an associated configuration space $\statespace^\agent$ with dimension $d^{\agent}\in \naturals$; for the sake of simplicity we assume $d^{\agent}$ to be the same for everyone.
Hence the joint configuration space of the agents is $\statespace=[0,1]^{d^{\agent}\card{\agents}}$ with a corresponding subset which is the \emph{obstacle region} $\obstaclespace$. 
Its complement defines the \emph{obstacle-free region} as the set closure of the difference $\freespace = \closure(\statespace / \obstaclespace)$. 
Each agent has associated an initial state $\state_{\init}^\agent \in \freespace$ and a goal region $\statespace_\goal^\agent\subseteq\statespace$.
Note that we borrow game theoretical notation to distinguish between quantities peculiar to a (sub)group of the agents and the joint quantities. A superscript specifies the agents of interest, no superscript denotes the joint quantities.
For instance, the configuration space of the $i$-th agent is $\statespace^\agent$, whereas $\statespace$ denotes the joint configuration space ($ = \Pi_{\agent\in\agents}\statespace^\agent$).

The \acrshort{ma-mpp} then boils down to finding a collision-free path for each of the agents to reach their respective goal from the given initial configuration.
More formally, let the path of the $i$-th agent be a function of the form $\mypath^\agent:[0,1] \rightarrow \statespace^\agent$. To avoid pathological cases, we consider only \emph{continuous} paths of \emph{bounded variation}~\cite{Karaman2011}.

\begin{definition}[\acrlong{ma-mpp}]
Let $\agents$ be a finite set of agents and the triplet $\left\langle \freespace, \{\statespace_\goal^\agent\}_{\agent \in \agents}, \{\state_\init^\agent\}_{\agent \in \agents} \right\rangle $. We define the \acrfull{ma-mpp}, in short $\problem(x_\init^\agents)$, as the problem of finding  $\jointpath =  \{ \mypath^{\agent} \}_{\agent \in \agents}$ such that:
\begin{enumerate}
    \item The path is \textit{feasible} for every agent
    \begin{equation}
        \mypath^{\agent}(0) = \state_\init^\agent \;\wedge\; \mypath^{\agent}(1) \in \closure(\statespace_\goal^\agent) \quad\forall \agent \in \agents;
    \end{equation}
    \item  The path is \textit{collision-free} w.r.t. the obstacles:
    \begin{equation}
        \mypath^{\agent}(\tau) \in \freespace \quad \forall \agent \in \agents \; \forall \tau \in [0,1];
    \end{equation}
    \item The path is \textit{collision-free} w.r.t. other agents:
\begin{equation}
    \coll(\mypath^{\agent}(\tau),  \mypath^\agenttwo(\tau))=\false \quad \forall_{\agent\neq\agenttwo}\agent, \agenttwo \in \agents ; \; \forall \tau \in [0,1].
\end{equation}
\end{enumerate}
If no such collection of paths exists, we report failure.
We denote the set of solution paths by $\sol(\problem(x_\init^\agents))$.
\end{definition}
Often, it is desirable to not only retrieve \emph{a} solution but to retrieve an optimal path according to a certain cost function, leading to the definition of:
\begin{definition}[Optimal \acrshort{ma-mpp}]\label{def:optimal_mampp}
Given a \acrfull{ma-mpp} $\problem(x_\init^\agents)$ with cost function 
$\cost: \Pi_{\agent \in \agents}\Sigma^{\agent} \to \reals_{\geq 0}$, a solution $\mypath^*$ is optimal if 
\begin{equation}
    \mypath^* \in \argmin_{\mypath \in \sol(\problem(x_\init^\agents))} \cost(\mypath).
\end{equation}
If no such path exists, we report failure. We denote the set of $\epsilon$-optimal paths by $\sol^{\epsilon}(\problem(x_\init^\agents))$.\\ That is, $\{\mypath \;|\; (\cost(\mypath)-\cost(\mypath^*))/\cost(\mypath^*)\leq \epsilon \}$) for $\epsilon\geq0$ . 
\end{definition}
In this work we consider as the cost function for the \acrshort{ma-mpp} the sum of the individual costs of the individual agents (i.e., the social cost), that is $\cost(\jointpath) = \sum_{\agent \in \agents} \cost^\agent(\jointpath)$.
We further take the individual cost function for an agent to be the individual \emph{path length} of the form $\cost^\agent: \Sigma^\agent \rightarrow \reals_{\geq 0}$.
This choice simplifies the presentation of this work restricting the interaction among the agents to be collisions only. Nevertheless, in more general cases, one can have coupling among the agents also at the cost level (e.g., a safety distance penalty)~\cite{Zanardi2021}. The concepts hereby presented would still be applicable (maybe with minor rework) as long as the \acrshort{ma-mpp} cost function is separable in the agents.

Finally, to ensure well-posedness of the \acrshortpl{sba}~\cite{Karaman2011}, we assume that~$\cost, \cost^\agent$ are \emph{monotonic}, i.e.,~$\forall \; \mypath, \mypath^\prime \in \pathspace,\; \cost(\mypath) \leq \cost(\mypath \cup \mypath^\prime)$ and \emph{bounded}, i.e., $\exists \; k_c \text{ such that } \cost(\mypath) \leq k_c TV(\mypath), \; \forall \mypath \in \pathspace$ where $k_c$ is a positive real constant, and $TV$ the total variation.
In the rest of this manuscript we will adopt the \emph{path length} as the cost function for the problem at hand. 

\subsection{\texorpdfstring{\acrlongpl{sba}}{Lg} for Motion Planning} \label{sec:SBAintro}
\acrfullpl{sba}~\cite[pg. 187]{LaValle2006} have emerged as a widely-used strategy for motion planning in robotics. \acrshortpl{sba} work by randomly sampling the configuration space of the agent and constructing a graph or roadmap, which expresses the connectivity and the feasible paths within the planning environment. 
In this manuscript, we consider \acrshortpl{sba} of the form presented in~\Cref{alg:sba}.
Note that for the scope of this work algorithms such as $\rrg,\,\prm,\,\prmstar,\,\rrt$ fit this scheme.
More sophisticated variations such as $\rrtstar$~\cite{Karaman2011}, $\fmtstar$~\cite{Janson2015FastDimensions}, $\bitstar$~\cite{Gammell2015BatchGraphs} would require redefining ad-hoc extra operations such as rewiring for factorization. 
We henceforth neglect them.
We identify four macro operations, namely $\getsample,\,\nearest,\,\collfree,\,\addedge$.
\subsubsection*{$\getsample$} gets the next sample to be processed from a pre-drawn pointset (e.g., $\prm$) or by drawing a new sample from $\freespace$ (e.g., $\rrg, \rrtstar$).
\subsubsection*{$\nearest$} the algorithm queries nodes from the graph in the neighborhood of the new random sample. 
Again, depending on the algorithm the query might return a single sample ($\rrt$), a set of neighbors within a certain radius ($\prm,\,\prmstar,\,\rrg$), or the \emph{k-nearest} vertices ($k$-$\prm$, $k$-$\prmstar$, $k$-$\rrg$).
\subsubsection*{$\collfree$} The function checks if the transition between $\state_{\rand}$ and $\state_{\near}$ is collision-free w.r.t. obstacles and the agents themselves for the multi-agent case. 
\subsubsection*{$\addedge$} Lastly if collision-free, $\state_\rand$ is added to the graph creating an edge with $\state_{\near}$. 
The edge of the graph is \emph{directed} or \emph{un-directed} depending on the algorithm.
\begin{assumption}\label{ass:sbas}
 For simplicity, in the remainder of this paper we refer to algorithms satisfying~\Cref{alg:sba} as \acrfullpl{sba}.
\end{assumption}
Among the algorithms satisfying~\Cref{ass:sbas}, we selected $\rrg$ and $\prmstar$ as the running example because of their optimality guarantees.
\begin{algorithm}
\caption{Sampling Based Algorithm}
\begin{algorithmic}[1]
\State $\graph.\addnode (\state_\init)$ 
\For{$i=0, \dots, N-1$}
\State $\state_{\rand} = \getsample(\freespace)$
\State $\nodeset_{\near} = \nearest(\state_{\rand})$
\For{$\state_{\near} \in \nodeset_{\near}$}
\If{$\collfree(\state_{\near}, \state_{\rand})$}
    \State $\graph.\addedge (\state_{\near}, \state_{\rand})$
\EndIf
\EndFor
\EndFor
\vspace{3px} \State \Return $\graph$
\end{algorithmic}
\label{alg:sba}
\end{algorithm}

\section{Factorization of \texorpdfstring{\acrshortpl{sba}}{Lg}}

We introduce the idea of factorization applied to \acrshortpl{sba}.
The result allows one to solve \acrshortpl{ma-mpp} in a ``hybrid'' fashion\footnote{In the sense of~\Cref{sec:relatedwork}.} more efficiently while preserving the completeness and optimality guarantees of \acrshortpl{sba} working in the joint space. This is achieved by leveraging a \emph{factorization} (decoupling) of the agents to independent lower dimensional search spaces where possible.
The key insight to understand the factorization idea is the observation that from certain joint configurations of the agents the remaining solutions can be searched independently. 
This is possible when the solution of the joint problem can be found by composing the solutions of smaller (sub)groups of agents that have been computed individually, pretending the other agents do not exist.
\begin{definition}[Independence of (Optimal) Motion Planning Problems]
\label{def:indep_of_sol}
Given a \acrshort{ma-mpp} $\problem(\state^\agents)$ and at least two disjoint set of agents $\agents', \agents'' \subseteq \agents$ we say that $\problem(\state^{\agents'})$ and $\problem(\state^{\agents''})$ are \textit{independent} for $\epsilon$-optimality, in symbols $\problem(\state^{\agents'}) \perp^{\epsilon} \problem(\state^{\agents''})$, if:
\begin{equation}\label{eq:mampp_indep}
\begin{aligned}
    &\mypath^{\agents} = \mypath^{\agents'} \cup \mypath^{\agents''}  \in \sol^{\epsilon}(\problem(\state^\agents)),\\
    &\forall \mypath^{\agents'} \in \sol^{\epsilon}(\problem(\state^{\agents'})), \;\forall \mypath^{\agents''} \in \sol^{\epsilon}(\problem(\state^{\agents''})).
\end{aligned}
\end{equation}
\end{definition}
Although general, the notion of independence given in~\Cref{def:indep_of_sol} can sometimes be impractical, as it requires to know the solutions of both the factorized and the joint problem to check for independence.
Nevertheless, we have found heuristics that are cheap to compute that imply independence, for example based on the future resource occupancy of the agents~\cite{ZanardiFactorization}. 
Therefore we consider a \emph{factorization heuristic} as a function that given two configurations tells if~\Cref{def:indep_of_sol} can be implied.
\begin{definition}(Factorization Heuristic) \label{def:independenceheuristic}
    Given a \acrshort{ma-mpp} $\problem(\state^{\agents}_{\init})$ and any two disjoint sets of agents $\agents', \agents'' \subset \agents$ we call a \emph{factorization heuristic} for $\epsilon$ optimality a function of the form
    \begin{equation}
        \facth:\statespace^{\agents'} \times \statespace^{\agents''} \to \mathtt{Bool},
    \end{equation}
    such that
    \begin{equation}
        \facth(\state^{\agents'}_{\init},\state^{\agents''}_{\init}) =
        1  \implies  \problem(\state^{\agents'}_{\init}) \perp^{\epsilon} \problem(\state^{\agents''}_{\init}).
    \end{equation}
\end{definition}
We recognise that the search of good factorization heuristics can be an entire research area \emph{per se}.
The specific type of the factorization heuristic in practice depends on the particular application, and often a sensible choice relies on additional insights and assumptions that can be made about the environment and the other agents (e.g., reachability analysis of cars driving in lanes, type of obstacles, behavior of other dynamic obstacles, crowdedness).

Different choices of the heuristic will determine how much a certain problem gets factorized and how much each iteration of~\Cref{alg:sba} will become computationally heavier.
On one extreme, if we were to directly evaluate~\Cref{def:indep_of_sol} we would obtain a ``perfect factorization'', i.e., the problem would be factorized as much as possible. This might in practice work well in environments that are not crowded, where the probability of computing the solutions for each agent individually and them being already collision-free is high, in other cases though, this might become computationally cumbersome.
On the other extreme, an heuristic that never factorizes would still be a valid function, clearly cheap to compute, but rather useless since it would return the original problem always working in the joint space of all the agents. Clearly there exists a whole Pareto front in between.
\subsection{Factorization Routine}\label{sec:factroutine}
To embed the factorization idea in \acrshortpl{sba}, we look for a function with signature: $\factorize:\statespace \to \partition(\statespace)$ where $\partition$ denotes the set of partitions\footnote{Recall that a partition of a set is a grouping of its elements into non-empty subsets, in such a way that every element is included in exactly one subset.}. For instance, consider a joint state of $3$ agents $\state = [\state^1,\state^2,\state^3]$, its partitions are $\{ [\state^1,\state^2,\state^3]\}$, $\{ [\state^1,\state^2],[\state^3] \}$, $\{ [\state^1],[\state^2,\state^3] \}$, $\{ [\state^1, \state^3],[\state^2] \}$, or $\{ [\state^1],[\state^2],[\state^3]\}$.
 
Despite in the remaining of this work we assume that a useful heuristic and $\factorize$ function are available, we hereby provide three concrete examples:
\begin{itemize}
    \item The first example is given by~\cite{ZanardiFactorization} using spatio-temporal occupancy of the robots.
    Given a state, one computes for each robot its forward reachability set, if two sets do not intersect the agents can be considered independent. This procedure applied to all the agents allows to build an independence graph\cite[Def. 7]{ZanardiFactorization} whose connected components represent the agents' partition.
    \item Generalizing the previous point, if additional information is available for the problem at hand one can factorize based on this.
    We will provide an example in~\Cref{sec:experiments} where we assume to know that, given a state, the optimal solution lives in a cone between us and the goal. We envision that similar ideas of informed search~\cite{Informed_rrtstar} can also apply.
    \item Learning to guess the partition of independent agents in combination with the definition itself. Given a joint state we try to factorize it according to a predicted partition of the agents.
\end{itemize}
\subsection{The Factorized \texorpdfstring{\acrlong{sba}}{Lg}}\label{sec:factosamplingbased}
In the following, we show how the factorization routine of~\Cref{sec:factroutine} can be integrated in~\Cref{alg:sba}.
Interestingly, the resulting motion graph is an hyper-graph that includes directed hyper-edges to go from one source to multiple destinations when agents get factorized (\Cref{fig:hypergraphedge}). This structure requires extra care in re-adapting some of the standard routines; for instance, the search of nearest neighbors of a factorized state.

A \emph{factorized \acrshort{sba}}, takes the form of~\Cref{alg:sbafact} with the following reformulation of the main operations:
\subsubsection{\texorpdfstring{$\getsample$}{Lg}}
This routine remains unchanged, the samples are still drawn from the joint space. Possibly they get factorized in the next step.
\subsubsection{\texorpdfstring{$\factorize$}{Lg}}
This operation factorizes the new sample by mapping it to one of its partitions as described in~\Cref{sec:factroutine}. 
We denote with the capital letter $\nodeset$ the sets of states.
For instance, $\nodeset_{\rand}$ is
\begin{equation}
     \begin{aligned}
         &\nodeset_{\rand} = \{\state^{\agents_i} : \problem(\state^{\agents_i}) \perp \problem(\state^{\agents_j}) \; \forall i \neq j \}.
     \end{aligned}
\end{equation}
\begin{figure}[tbh]
\begin{subfigure}{.4\linewidth}
  \centering
  \raisebox{6mm}{\resizebox{\linewidth}{!}{\tikzset{every picture/.style={line width=0.75pt}} %

\begin{tikzpicture}[x=0.75pt,y=0.75pt,yscale=-1,xscale=1]
\draw  [dash pattern={on 4.5pt off 4.5pt}] (115,40) -- (208,40) -- (208,142) -- (115,142) -- cycle ;
\draw  [fill={\oneplayercolor}  ,fill opacity=1 ] (142.5,115.25) .. controls (142.5,105.17) and (150.67,97) .. (160.75,97) .. controls (170.83,97) and (179,105.17) .. (179,115.25) .. controls (179,125.33) and (170.83,133.5) .. (160.75,133.5) .. controls (150.67,133.5) and (142.5,125.33) .. (142.5,115.25) -- cycle ;
\draw  [fill={\oneplayercolor}  ,fill opacity=1 ] (34.5,134) .. controls (34.5,125.16) and (41.66,118) .. (50.5,118) .. controls (59.34,118) and (66.5,125.16) .. (66.5,134) .. controls (66.5,142.84) and (59.34,150) .. (50.5,150) .. controls (41.66,150) and (34.5,142.84) .. (34.5,134) -- cycle ;
\draw  [fill={\threeplayercolor}  ,fill opacity=0.5 ] (12,89.5) .. controls (12,78.18) and (29.24,69) .. (50.5,69) .. controls (71.76,69) and (89,78.18) .. (89,89.5) .. controls (89,100.82) and (71.76,110) .. (50.5,110) .. controls (29.24,110) and (12,100.82) .. (12,89.5) -- cycle ;
\draw  [fill={\twoplayercolor}  ,fill opacity=0.5 ] (18,44) .. controls (18,32.95) and (33.67,24) .. (53,24) .. controls (72.33,24) and (88,32.95) .. (88,44) .. controls (88,55.05) and (72.33,64) .. (53,64) .. controls (33.67,64) and (18,55.05) .. (18,44) -- cycle ;
\draw  [fill={\twoplayercolor}  ,fill opacity=0.5 ] (125,64) .. controls (125,52.95) and (140.67,44) .. (160,44) .. controls (179.33,44) and (195,52.95) .. (195,64) .. controls (195,75.05) and (179.33,84) .. (160,84) .. controls (140.67,84) and (125,75.05) .. (125,64) -- cycle ;
\draw [color={\threeplayercolor}  ,draw opacity=1 ]   (89,89.5) .. controls (113.98,89.96) and (119.27,85.72) .. (129.42,80.33) ;
\draw [shift={(132,79)}, rotate = 153.43] [fill={\threeplayercolor}  ,fill opacity=1 ][line width=0.08]  [draw opacity=0] (10.72,-5.15) -- (0,0) -- (10.72,5.15) -- (7.12,0) -- cycle    ;
\draw [color={\threeplayercolor}  ,draw opacity=1 ]   (89,89.5) .. controls (115.06,89.98) and (121.55,88.14) .. (141.75,103.28) ;
\draw [shift={(144,105)}, rotate = 217.69] [fill={\threeplayercolor}  ,fill opacity=1 ][line width=0.08]  [draw opacity=0] (10.72,-5.15) -- (0,0) -- (10.72,5.15) -- (7.12,0) -- cycle    ;
\draw [color={\oneplayercolor}  ,draw opacity=1 ]   (66.5,134) .. controls (99,133.03) and (112.2,128.3) .. (139.89,116.38) ;
\draw [shift={(142.5,115.25)}, rotate = 156.63] [fill={\oneplayercolor}  ,fill opacity=1 ][line width=0.08]  [draw opacity=0] (10.72,-5.15) -- (0,0) -- (10.72,5.15) -- (7.12,0) -- cycle    ;
\draw [color={\twoplayercolor}  ,draw opacity=1 ]   (88,44) .. controls (103.98,44) and (114.66,46.65) .. (128.34,51.12) ;
\draw [shift={(131,52)}, rotate = 198.43] [fill={\twoplayercolor}  ,fill opacity=1 ][line width=0.08]  [draw opacity=0] (10.72,-5.15) -- (0,0) -- (10.72,5.15) -- (7.12,0) -- cycle    ;
\draw  [dash pattern={on 0.84pt off 2.51pt}] (3,21) -- (95,21) -- (95,159) -- (3,159) -- cycle ;

\draw (183,17) node [anchor=north west][inner sep=0.75pt]   [align=left] {$\displaystyle \nodeset_{\rand}$};
\draw (32,83) node [anchor=north west][inner sep=0.75pt]   [align=left] {$\state^{\boldsymbol{1,2,3}}_\near$};
\draw (37,128) node [anchor=north west][inner sep=0.75pt]   [align=left] {$\state^{\boldsymbol{3}}_\near$};
\draw (35,38) node [anchor=north west][inner sep=0.75pt]   [align=left] {$\state^{\boldsymbol{1,2}}_\near$};
\draw (143,107) node [anchor=north west][inner sep=0.75pt]   [align=left] {$\state^{\boldsymbol{3}}_\rand$};
\draw (140,55) node [anchor=north west][inner sep=0.75pt]   [align=left] {$\state^{\boldsymbol{1,2}}_\rand$};
\draw (61,-3) node [anchor=north west][inner sep=0.75pt]   [align=left] {$\nodeset_{\near}$};

\end{tikzpicture}}}
  \caption{}
  \label{fig:powersetconnection}
\end{subfigure}
\begin{subfigure}{.59\linewidth}
  \centering
  \includegraphics[width = \linewidth]{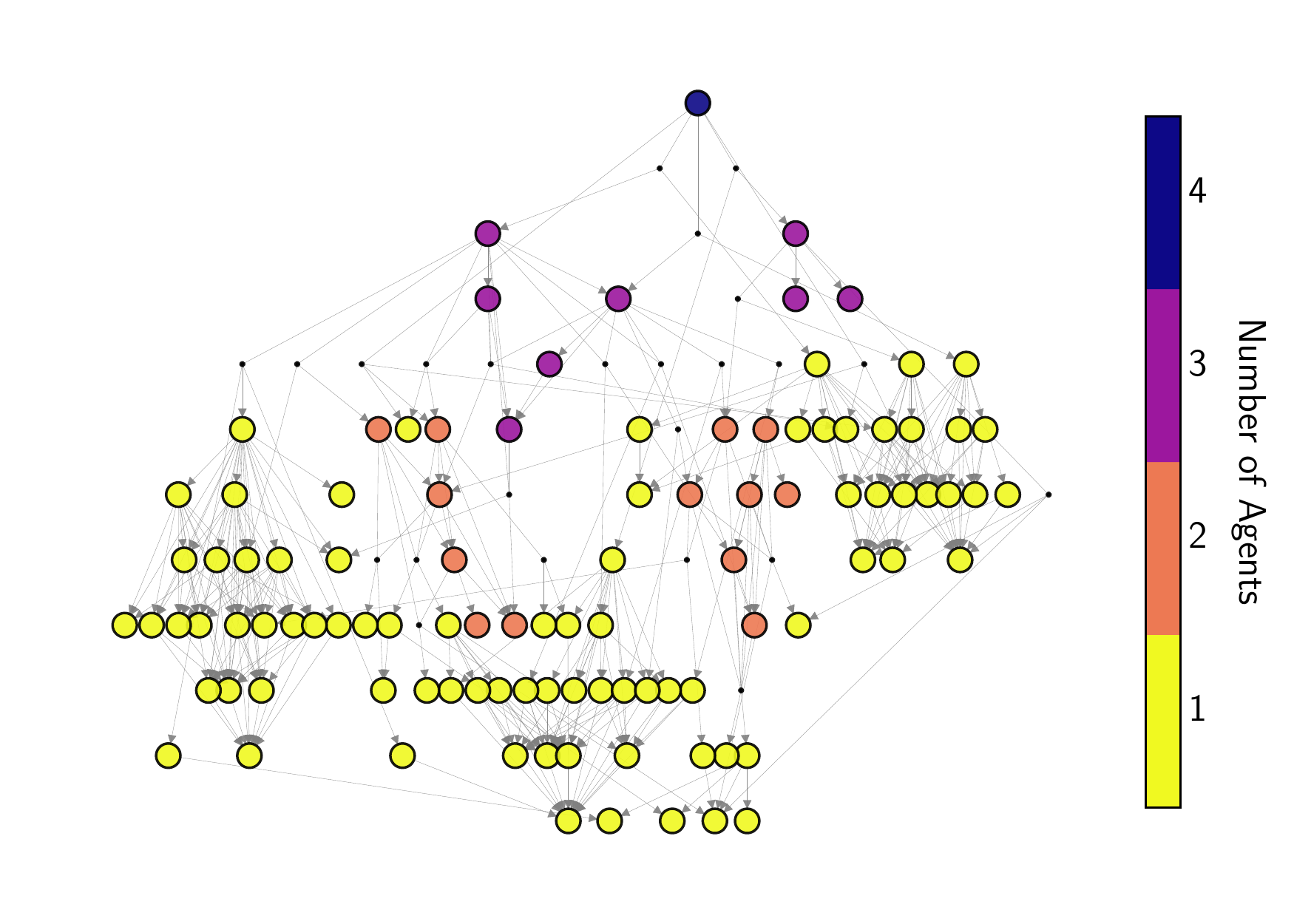}
  \caption{}
  \label{fig:hypergraphedge}
\end{subfigure}%
\caption{In~\Cref{fig:powersetconnection} a random sample $\state_\rand$ gets factorized into $\nodeset_{\rand} =\{[\state_\rand^1, \state_\rand^2],[\state_\rand^3] \}$. Importantly, the nodes of $\nodeset_{\rand}$ can be reached from neighbouring nodes of two types: either nodes with the same agents, or from nodes containing a (partial) union of the agents. The latter are indeed the transition across which agents get factorized. This structure requires to search the neighbors in $\powerset(\nodeset_{\rand})$. 
\Cref{fig:hypergraphedge} shows an example of the full graph resulting by running the factorized version of $\rrg$ (\Cref{alg:sbafact}) in a $4$-agent problem. 
The initial state is 4-dimensional but quickly gets decoupled into a 3-agent node and a 1-agent node, the graph eventually gets factorized into four single-agent components. The decoupling of agents during the transitions are depicted with a small black dot.
}
\end{figure}
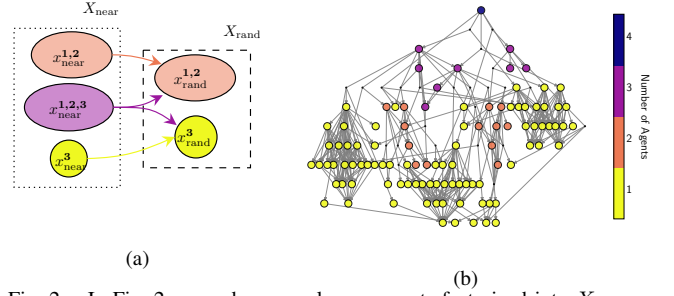
\begin{algorithm}[!htbp]
\caption{FactSBA}
\begin{algorithmic}[1]
\State $\nodeset_{\init} = \factorize(\state_{\init})$
\State $\hypergraph.\addnode (\nodeset_\init)$
\For{$i=0, \dots, N-1$}
\State $\state_{\rand} = \getsample(\freespace)$
\State $\nodeset_{\rand} = \factorize(\state_{\rand})$
\For {$\widehat{\state}_{\rand} \in \powerset(\nodeset_{\rand}) $}
\State $\nodeset_{\near} = \factnearest(\widehat{\state}_{\rand})$
\For{$\state_{\near} \in \nodeset_{\near}  $}
\If{$\collfree(\state_{\near}, \widehat{\state}_{\rand})$}
    \State $\hypergraph.\addedge(
 \state_{\near},\nodeset^{\agents(\widehat{\state}_{\rand})}_{\rand})$
\EndIf
\EndFor
\EndFor
\EndFor
\vspace{2px} \State \Return $\hypergraph$
\end{algorithmic}
\label{alg:sbafact}
\end{algorithm}
\subsubsection{\texorpdfstring{$\nearest$}{Lg}}
To understand the difference between this operation and the one in~\Cref{alg:sba}, it is crucial to note that the nodes in $\nodeset_{\rand}$ can be reached both by neighbors containing the same set of agents (e.g., $\{ \state^3\} \to \{(\state^3)^+ \}$, as well as by nodes containing a superset of the agents (e.g., $\{ (\state^3)^+\}$ can be reached by a node $\{ \state^{1,2,3} \}$ that gets factorized in the transition, as per~\Cref{fig:powersetconnection}).
Generalizing, all the possible combinations of agents' partitions contained in $\nodeset_{\rand}$ correspond to the powerset $\powerset(\cdot)$ of $\nodeset_{\rand}$, which yields both the single elements and their combinations.
Therefore, in line $6$-$7$, for each element of the powerset we define the $\factnearest$ operation which actually looks for the nearest neighbors containing the same set of agents of $\widehat{\state}_{\rand}$. Importantly, note that $\widehat{\state}_{\rand}$ needs to be treated as a joint state.

In general, $\nodeset_\near$ is a again a set, usually determined by the connection radius or by the $k$-nearest nodes. As observed in~\cite{Karaman2011} the optimal connection radius is a decreasing function of the dimension $d$ of the search space, which in the factorized case varies from node to node. 
One could potentially adapt the connection radius online to fit the dimension of the input state, but also keeping this radius as a function of the dimension of the largest joint search space suffice for convergence.
Finally, from a computational perspective, the number of nearest neighbors queries per iteration increases on average since we check for every element of the powerset. Nevertheless these queries only regard a smaller subset of the whole graph. 
In our case study we found this not to burden on the average computation time per iteration.
\subsubsection{\texorpdfstring{$\collfree$}{Lg}} 
This routine does not change substantially from~\Cref{alg:sba}. We highlight only that the transition has to be considered joint among the agents in $\widehat{\state}_{\rand}$ even when the edge will be decoupling the agents. This guarantees that the decoupling transitions are collision-free.
\subsubsection{\texorpdfstring{$\addedge$}{Lg}} 
Differently from~\Cref{alg:sba}, the $\addedge$ operation of the factorized algorithm adds an \emph{hyperedge}.
The source is always a single node $\state_\near$, while the destination can be a (sub)set of nodes in $\nodeset_{\rand}$.
We denote by $\nodeset^{\agents(\widehat{\state}_{\rand})}_{\rand}$ the subset of $\nodeset_{\rand}$ which contains the same agents of $\widehat{\state}_{\rand}$. In other words, $\nodeset^{\agents(\widehat{\state}_{\rand})}_{\rand}$ is the factorized version of $\widehat{\state}_{\rand}$, which is used as a joint state for nearest neighbor search and collision check. 
Therefore, the hyperedge will result in a standard edge if $\nodeset^{\agents(\widehat{\state}_{\rand})}_{\rand}$ contains only the state $\widehat{\state}_{\rand}$, and a splitting edge if $\nodeset^{\agents(\widehat{\state}_{\rand})}_{\rand}$ contains more states.
The different cases are depicted in~\Cref{fig:powersetconnection}.

Importantly, we highlight that the choice of certain factorization heuristics may induce additional ``coherency'' constraints for new edges. For instance, if one uses ``future resources'' of the search space as heuristic, the destination nodes needs to belong to the future resources of the origin to preserve the decoupling. 
Finally, as for the non-factorized version, the cost of the agent transition is associated to the edge.
For this work, a splitting edge considers the sum of the costs of individual transitions.  
\subsection{Properties of the Factorized \texorpdfstring{\acrshortpl{sba}}{Lg}}
In the following, we show that the factorized counterpart of an asymptotically optimal \acrshortpl{sba} preserves its optimality properties, i.e., converges to an optimal solution as the number of samples goes to infinity.
\begin{proposition}(Convergence of Factorized \acrshortpl{sba}) 
The factorized counterpart of an asymptotically optimal \acrshortpl{sba} preserves its optimality properties, i.e., converges to an optimal solution as the number of samples goes to infinity.
\end{proposition}
\begin{proof}
The key intuition resides in observing that the path found by Fact\acrshort{sba} can be seen as the composition of solutions found by standard \acrshortpl{sba} algorithms running on equal or lower dimensional search spaces.
First, observe that in~\Cref{alg:sbafact}, when $\widehat{\state}_{\rand}$ coincides with $\nodeset^{\agents(\widehat{\state}_{\rand})}_{\rand}$--i.e., $\nodeset^{\agents(\widehat{\state}_{\rand})}_{\rand}$ is not factorized--the whole procedure ($\nearest,\,\addedge$) coincides to the one in~\Cref{alg:sba}. This is exclusive to the subgraph of $\hypergraph$ with the nodes relative to the (joint) state with agents $\agents(\widehat{\state}_{\rand})$.
Indeed, if we never factorize any state, \Cref{alg:sbafact} is the same as~\Cref{alg:sba}.

Now consider the following: the best path found by~\Cref{alg:sbafact} after $n$ iterations has by construction only a finite number (if any) of splitting hyper-edges (e.g.,~\Cref{fig:hypergraphedge}). These are finite since we have a finite number of players that we can factorize.
Hence, consider $\factorize$ built by any heuristic satisfying~\Cref{def:independenceheuristic} for a given $\epsilon \geq 0$. 
By construction, a non-trivial $\factorize$ decomposes portions of the joint space into their ``marginals'', i.e., the projections relative only to a specific subset of players; we call these subsets \emph{blocks}, as reminiscence of the blocks of a partition.
Then~\Cref{alg:sbafact} runs~\Cref{alg:sba} for each block finding the optimal path from the initial node of the block to the subsequent partition (or to the goal). Note that for the first block its ``initial node'' is fixed ($\state_\init$).\looseness=-1 

Finally, if we let \Cref{alg:sba} run long enough, for each block, the found path almost surely will be at least $\epsilon$-optimal, hence satisfies the decoupling condition. 
Since this holds for any block, the overall path is also $\epsilon$-optimal since the product of factorized $\epsilon$-optimal paths as defined in~\Cref{def:optimal_mampp} is again $\epsilon$-optimal for the joint problem (\Cref{thm:epsiloncomposition}), and the same holds for the concatenation of $\epsilon$-optimal paths (\Cref{thm:epsilonunion}).
\end{proof}

The proof highlights an important aspect. If we factorize according to an $\epsilon$-optimality, the decoupling ensures feasibility only when we find solutions at least $\epsilon$-optimal in the sub-problems. 
Indeed we observe that while a large epsilon would allow to factorize less, it allows to find guaranteed-to-be-decoupled solutions more quickly. But choosing and estimating online $\epsilon$ might be impractical.

Nonetheless there exist ``workarounds'' to enforce that any found solution is feasible (and not only the one at least $\epsilon$-optimal). For example, one can use heuristics that associate a set of states to each node and factorize if these sets are not intersecting.
Then add a connection (line 10 of~\Cref{alg:sbafact}) only if $\nodeset^{\agents(\widehat{\state}_{\rand})}_{\rand}$ belongs to this set for $\state_{\near}$ and these sets are forward invariant, each contained in the set of the parent, we are guaranteed that any feasible path found in that block is already validly decoupled (e.g., \cite{ZanardiFactorization}).
\section{Sample Gain of $\prmstar$ via Factorization}
In the following, we introduce a \emph{sample complexity gain}. Informally defined as the ratio between the minimum sufficient number of samples required to achieve a certain optimality with a certain probability with and without factorization. 
The gain clearly will depend on an intrinsic ``factorizability'' of the problem at hand, i.e., for which fraction of the joint configuration space we can factorize. Importantly, it measures the improvement over the slow convergence rates of \acrshortpl{sba} working in the joint space which is often at least exponential in the dimension of the search space~\cite[pg. 324]{LaValle2006}.
\begin{definition}[Sample Complexity Gain via Factorization]\label{def:factgain}
Let $\problem(\state_\init^\agents)$ be an optimal \acrshort{ma-mpp} defined in~\Cref{def:optimal_mampp}.
Then fix an arbitrary desired optimality $\epsilon>0$ and a probability level $\pbar\in[0,1]$ such that $\prob[(\cost(\mypath^\alg)-\cost(\mypath^*))/\cost(\mypath^*)\leq \epsilon]\geq \pbar$.
We define the \emph{sample complexity gain of factorization} as the percentage of the spared samples saved by the factorized \acrshort{sba} to achieve $\epsilon$-optimality with probability at least $\pbar$. This quantity reads
\begin{equation}\label{eq:sampleratio}
    \gfact \coloneqq 1-\frac{ \minN^{\fact}_{\epsilon,\pbar}}{\minN_{\epsilon,\pbar}}.
\end{equation}
\end{definition}
In the following we derive~\Cref{def:factgain} for the specific case of $\prmstar$. 
First recall that for most of \acrshortpl{sba}, requiring a certain optimality is equivalent to achieving a certain low dispersion of the sample sequence (\Cref{def:dispersion})\footnote{with a little abuse of notation here we use the term ``sampled sequences'' and ``pointsets'' interchangeably.}. 
Hence we consider:
\begin{definition}[Minimal Samples for Desired Dispersion]\label{def:suffsamples}
Let $\pointset$ be a dense (in the limit) sequence on $[0,1]^d$. Fix $\dispbar,\pbar \in (0,1]$ to be respectively a desired dispersion and probability level. We denote by $\minN_{\dispbar,\pbar}\in\naturals$ the minimum number of samples of $\pointset$ such that $\prob(\disp(\pointset)<\dispbar)\geq\pbar$. 
\end{definition}
For $\ell_\infty$-dispersion and uniform sampling, we approximate~\Cref{def:suffsamples} taking the smallest $N$ satisfying~\Cref{th:suffsamples}.
We now introduce the definition of the \emph{Factorization Factor} which quantifies how much a certain \acrshort{ma-mpp} can be factorized.
\begin{definition}[Factorization Factor]
    Given a \acrshort{ma-mpp} $\problem(x_\init^\agents)$ defined over the (joint) freespace $\freespace$, then for a possible partition of the agents $\partition$ we define its \emph{factorization factor} as the fraction of $\freespace$ that can be factorized according to $\partition$ and not more--denoted as $\freespace^\partition$. We denote it as 
    \begin{equation}
        \factratio^{\partition} \coloneqq \frac{\lebesgue(\freespace^\partition)}{\lebesgue(\freespace)}.
    \end{equation}
\end{definition}

\begin{proposition}\label{thm:fact_gain}(Sample Complexity Gain for $\factprm$) 
Given an optimal \acrshort{ma-mpp} $\problem(\state_\init^\agents)$ as in~\Cref{def:factgain}
an arbitrary desired optimality $\epsilon \geq 0$ with probability level $\pbar \in [0,1]$.
Adopting uniform sampling and $\prmstar$ as the \acrshort{sba}, assuming for simplicity that a fraction $\factratio$ of $\freespace=[0,1]^d$ is ``fully factorizable''\footnote{the partition where all the agents are factorized into individual sets.}, then the \emph{sample complexity gain via factorization} reads:
\begin{equation}\label{eq:prmgain}
    \gfact_{\factprm} =\factratio - \frac{(1-\factratio)\log{\left(1- \factratio\right)}}{\log{\left(\frac{\lebesgue}{\dispbar^{d^i\card{\agents}}(1-\pbar)}\right)}}  + \bigo(\dispbar^{d^i(\card{\agents}-1)}).
\end{equation}
\end{proposition}

\begin{proof}
    Let $\factprm $ represent the \emph{factorized} counterpart of $\prmstar$.
    Let $\dispbar = \disp_{\infty}(\pointset)$ be the sufficient dispersion such that the optimality requirement $\epsilon$ is satisfied according to \cite[Theorem 2]{Janson2018DeterministicPerformance}.
    Note that $\minN_{\dispbar,\pbar}$ can be obtained from~\Cref{th:suffsamples}.
    Therefore the \emph{minimum} number of samples we need for $\prmstar$ working in the joint space is:
    \begin{equation} \label{eq:numjoint}
    \minN^{\prmstar}_{\dispbar,\pbar} = \frac{\lebesgue(\freespace)}{\dispbar^{d^i \card{\agents}}} \log{\left( \frac{\lebesgue(\freespace)}{\dispbar^{d^i \card{\agents} }(1-\pbar)}\right)}.
    \end{equation}
    By assumption, the \emph{factorized} algorithm $\factprm$ is able to perfectly decouple the agents for samples that belong to the ``fully factorizable`` region. Then we can separate the search space of $\factprm$ in two: 
    First, the ``non factorizable'' subregion of the joint space, which has Lebesgue measure $(1-\factratio)\lebesgue(\freespace)$ and dimension $d^i\card{\agents}$. 
    Second, the factorized subregion which corresponds to $\card{\agents}$ problems each with dimensionality $d^i$ and Lebesgue measure $(\factratio \lebesgue(\freespace))^{\frac{1}{\card{\agents}}}$. 
    Note, that to ensure that the optimality requirement is satisfied with probability $\pbar$ overall, we need to strengthen the probability requirement of each subproblem, so that their simultaneous occurrence happens with probability $\pbar$. For each sub-problem we define $\pbar^i \coloneqq \pbar^{\frac{1} {\card{\agents}}}$ as the strengthened probability level.
    
    To obtain the minimum number of samples we sum the samples required by each of the subproblems, but because when a sample gets factorized it provides already $\card{\agents}$ lower dimensional samples, we have:
    \begin{equation}\label{eq:samplecomposition}
         \minN^{\factprm}_{\dispbar,\pbar} =  \minN^{\mathrm{joint}}_{\dispbar,\pbar} +  \minN^{\mathrm{i}}_{\dispbar,\pbar^i},
    \end{equation}
    where by $ \minN^{\mathrm{joint}}_{\dispbar,\pbar}$ and $\minN^{\mathrm{i}}_{\dispbar,\pbar^i}$ correspond to~\Cref{def:suffsamples} for the ``joint region'' and ``the single agent'' respectively.  
    Note indeed that~\eqref{eq:samplecomposition} preserves the overall optimality, since each of the subproblems is guaranteed to satisfy the optimality requirement and so is their composition according to~\Cref{thm:epsiloncomposition} and~\Cref{thm:epsilonunion}. 
    By applying~\Cref{th:suffsamples} and using $\lebesgue$ as shorthand for $\lebesgue(\freespace)$ we obtain:
    \begin{equation}\label{eq:numfact}
    \begin{aligned}
        \minN^{\factprm}_{\dispbar,\pbar} &=  \frac{(1- \factratio)\lebesgue}{\dispbar^{d^i \card{\agents}}} \log{\left( \frac{(1- \factratio)\lebesgue}{\dispbar^{d^i \card{\agents} }(1-\pbar)}\right)} \\
        &+ \frac{(\factratio\lebesgue)^{\frac{1}{\card{\agents}}}}{\dispbar^{d^i}} \log{\left( \frac{(\factratio\lebesgue)^{\frac{1}{\card{\agents}}}}{\dispbar^{d^i}(1-\pbar^{\frac{1} {\card{\agents}}})}\right)}
    \end{aligned}
    \end{equation}
    Lastly substituting \eqref{eq:numfact} and \eqref{eq:numjoint} into \eqref{eq:sampleratio} we have $\frac{ \minN^{\fact}_{\epsilon,\pbar}}{\minN_{\epsilon,\pbar}}=$
    \begin{equation}
            \frac{\frac{(1- \factratio)\lebesgue}{\dispbar^{d^i \card{\agents}}} \log{\left( \frac{(1- \factratio)\lebesgue}{\dispbar^{d^i \card{\agents} }(1-\pbar)}\right)}+
            \frac{(\factratio\lebesgue)^{\frac{1}{\card{\agents}}}}{\dispbar^{d^i}} \log{\left( \frac{(\factratio\lebesgue)^{\frac{1}{\card{\agents}}}}{\dispbar^{d^i}(1-\pbar^{\frac{1} {\card{\agents}}})}\right)}
            }{\frac{\lebesgue}{\dispbar^{d^i \card{\agents}}} \log{\left( \frac{\lebesgue}{\dispbar^{d^i \card{\agents} }(1-\pbar)}\right)}}, %
    \end{equation} 
    which can be brought to~\eqref{eq:prmgain}.
    \end{proof}
\begin{figure}
\begin{subfigure}{.49\linewidth}
    \centering
    \includegraphics[width = \linewidth]{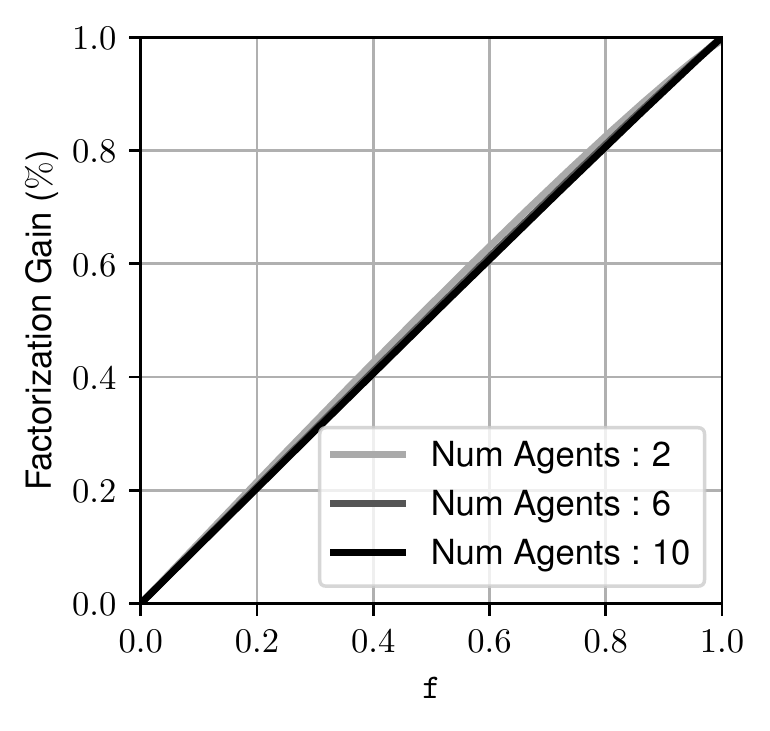}
    \caption{}
        \label{fig:FactorizationGainF}
\end{subfigure}
\begin{subfigure}{.49\linewidth}
    \centering
    \includegraphics[width = \linewidth]{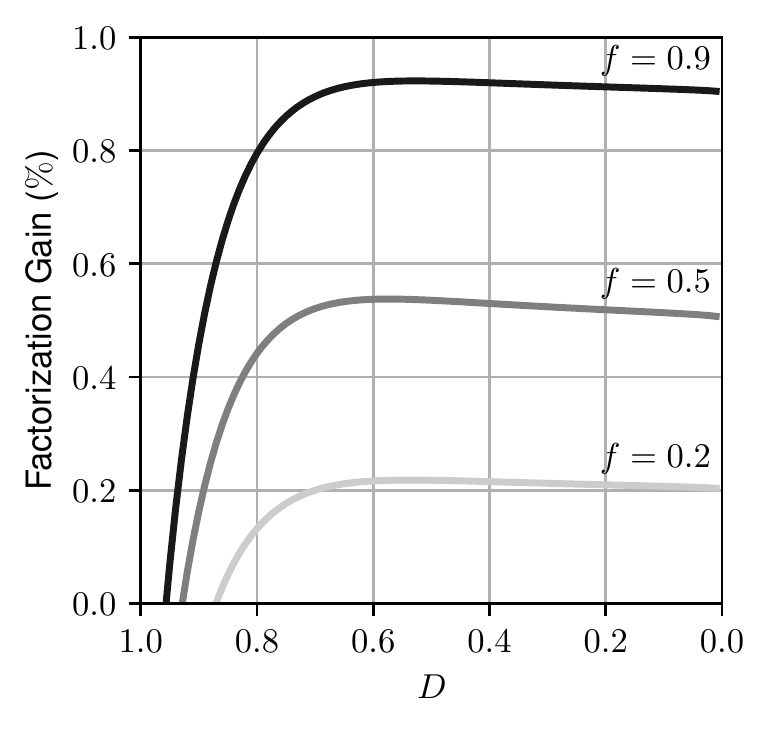}
    \caption{}
    \label{fig:FactorizationGainH}
\end{subfigure}
\caption{\Cref{fig:FactorizationGainF} shows how the factorization gain is indeed almost linear w.r.t. the ``factorizability`` of the problem. Clearly while the gain is almost invariant to the number of players, clearly the absolute number of samples is not.
In \Cref{fig:FactorizationGainH} we show the factorization gain as a function of the require optimality (hence lower $\ell_\infty$-dispersion). One can observe that $\gfact_{\factprm} \rightarrow \factratio$ as $\dispbar\rightarrow 0$.
In general we found these plots to be little sensitive to different choices of parameters; the ones displayed fix $\pbar=.7$, $d^i=2$, $\dispbar=.7$ (\Cref{fig:FactorizationGainF}), and $\card{\agents}=5$ (\Cref{fig:FactorizationGainF}).
}
\end{figure}

\section{Experiments}\label{sec:experiments}
We benchmark the factorization of \acrshort{sba} on $\rrg$\cite{Karaman2011}; we name its factorized counterpart $\factrrg$. 
We chose $\rrg$ as the running example since we deem it to be the simplest among the \acrshortpl{sba} with optimality and anytime properties without requiring rewiring or any sophistication on top of~\Cref{alg:sba}.

\begin{figure}[htb!]
    \centering
    \begin{subfigure}[t]{.49\linewidth}
           \centering
            \includegraphics[width = \linewidth]{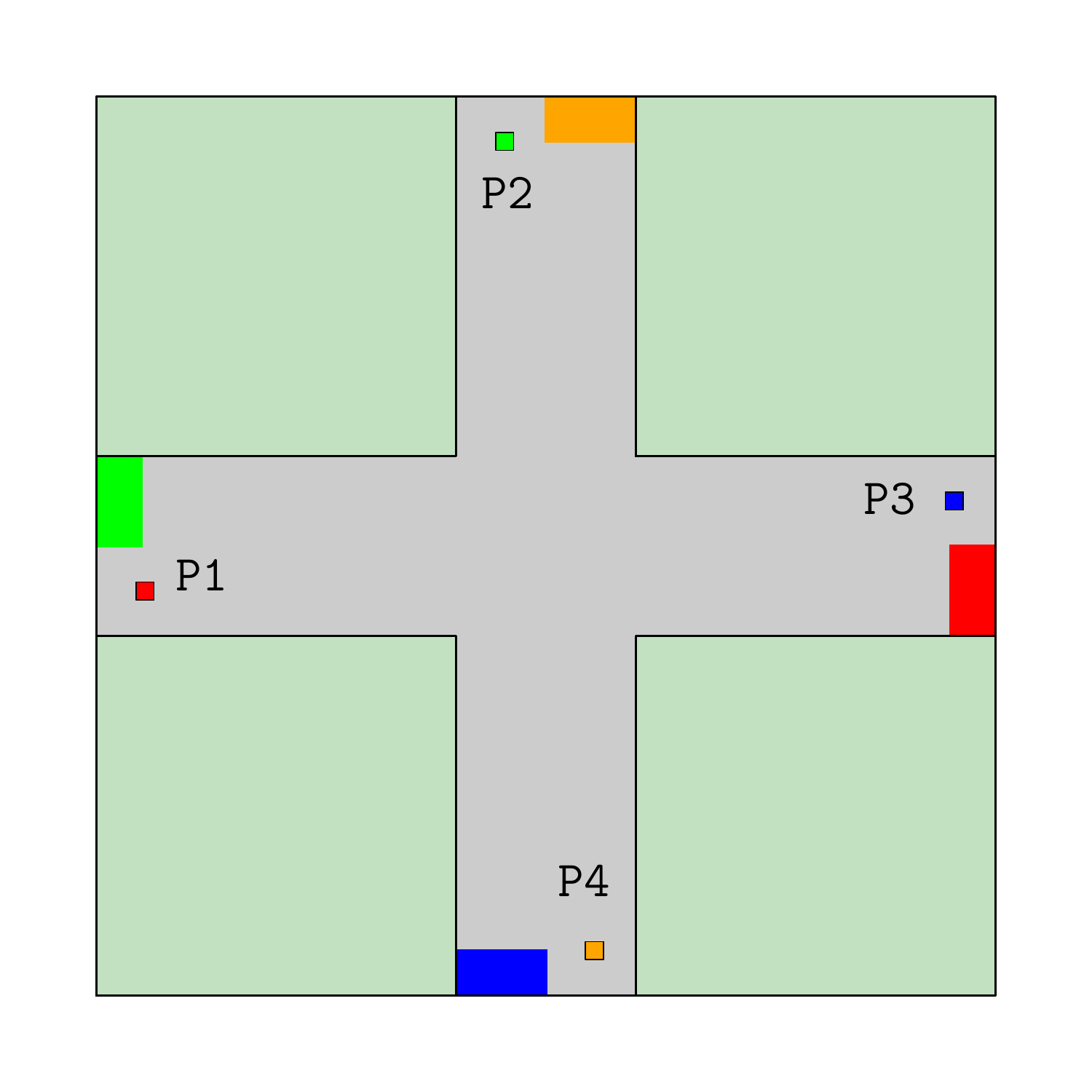}
            \caption{Scenario}
           \label{fig:scenario}
    \end{subfigure}
    \begin{subfigure}[t]{.49\linewidth}
           \centering
            \includegraphics[width = \linewidth]{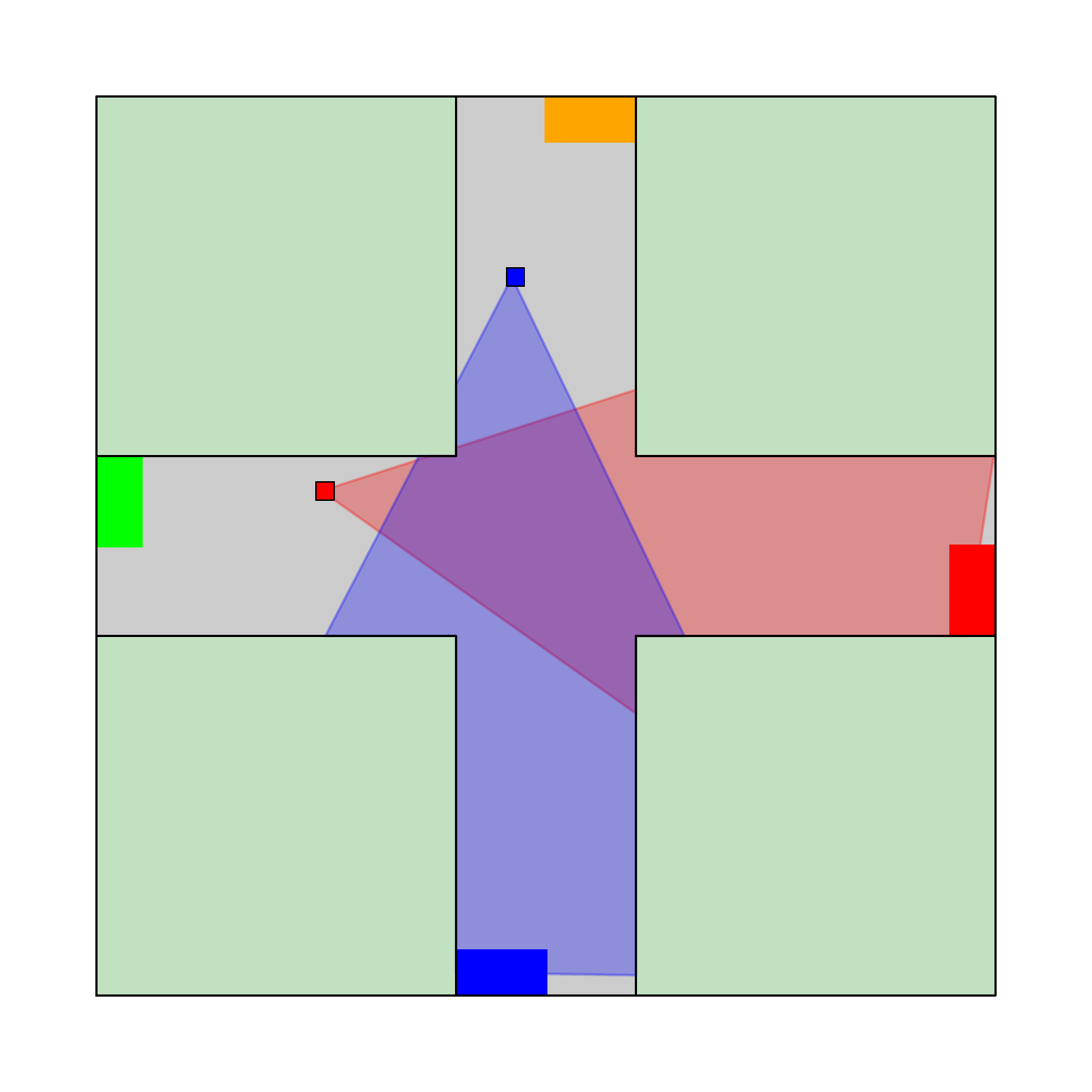}
           \caption{Future Resources}
           \label{fig:futureres}
    \end{subfigure}
    \medskip
    \caption{In~\Cref{fig:scenario} we show the evaluation scenario for the $4$-agent case. The $2$ and $3$-agent cases are obtained by taking players in order ($\mathtt{P1},\mathtt{P2}$), $3$ ($\mathtt{P1},\mathtt{P2},\mathtt{P3}$), and $4$ ($\mathtt{P1},\mathtt{P2},\mathtt{P3},\mathtt{P4}$). In~\Cref{fig:futureres} we show an instance of the \emph{future-resources} computation used in the experiments.}
\end{figure}

The evaluation scenario is depicted in~\Cref{fig:scenario}, we consider three cases with $2$, $3$, and $4$ agents respectively.
The agents are assigned an initial condition at the end of the corridors and a goal region in one of the opposite ones. The central intersection forces the agents to ``stay coupled'' for most of the space. 
The agents are \emph{holonomic} with no dynamic constraints. The state of each agent consists in its position on the 2D plane, thus $\state\in\reals^2$; the controls inputs are the velocities $[\speed_x, \speed_y]\in\reals^2$ in the $x$ and $y$ directions.
Clearly extensions to more complicated agents with dynamic constraints and scenarios are possible, but out of scope for this manuscript.
\begin{figure} 
\centering
    \includegraphics[width = \linewidth]{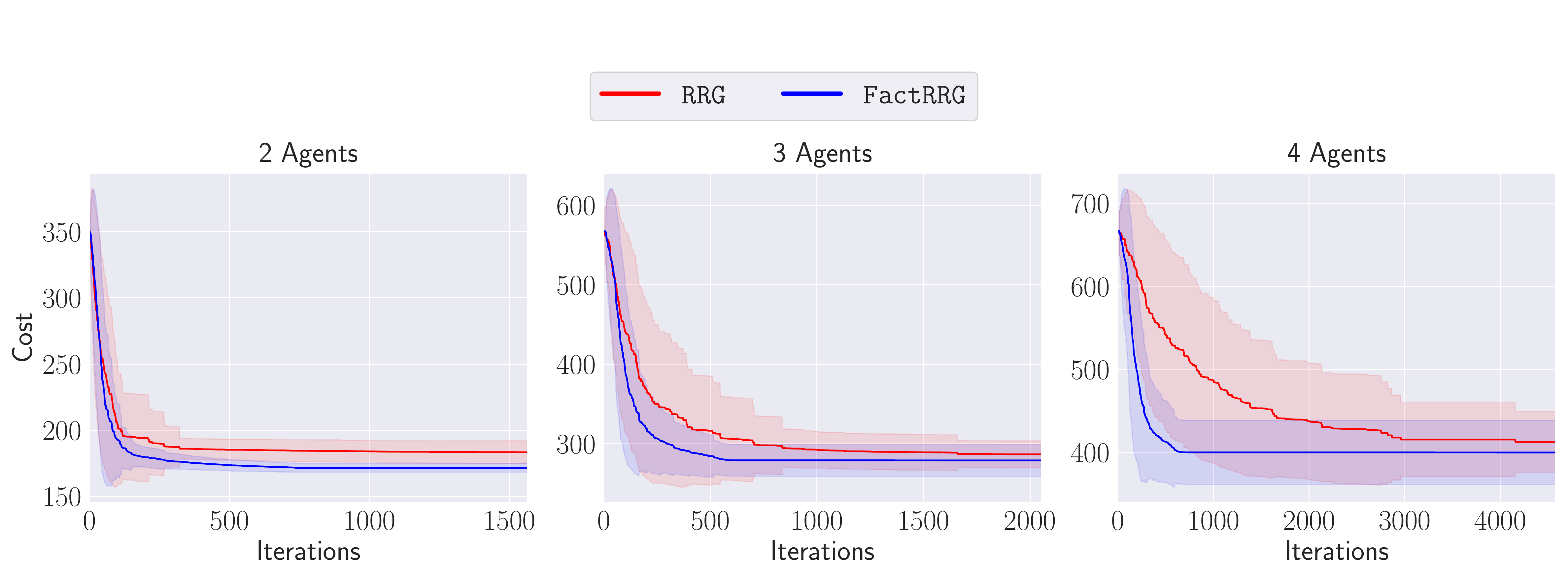}
    \caption{ 
    Cost as a function of number of iterations for $\rrg$ algorithm and its factorized counterpart $\factrrg$.
    The algorithms have been run with the same parameters and random sequences one after the other, for $100$ repeated trials.
    Both the average cost and the relative $1$-standard deviation band show that as we increase the number of players the factorized counterpart becomes more effective.
}
\label{fig:costevolution}
\end{figure}
\paragraph*{Implementation details}
The implementation of $\rrg$ and $\factrrg$ reflect~\Cref{alg:sba} and~\Cref{alg:sbafact} respectively.
We set the \emph{connection radius} to be optimal for $\rrg$~\cite{Karaman2011} in both versions. This takes the form $r_{\rrg} = \min{(\gamma_{\rrg} \left(\frac{\log{(n)}}{n}\right)^{\frac 1 d}, \eta_{\rrg})}$, with $n$ being the number of nodes and the user defined parameters set to $\gamma_{\rrg} = 100$,  $\eta_{\rrg} = 100$. 
Furthermore, to speed up convergence, the sampling procedure is set to return a point within the goal region with probability $0.1$ in both algorithms.

The two algorithms were run in succession for each trial with the same random seed. This to ensure the same sampled sequence for both algorithms. 
For the experiments, we used as stopping criterion a fixed finite number of graph nodes; more precisely, we stop the execution when a solution is found and the number of nodes in the respective graph surpasses $1k$.

The \emph{factorization heuristic} adopted is illustrated in~\Cref{fig:futureres}.
It relies on the additional assumptions that we make on the environment, namely, that for each agent there exist a solution path in a cone between itself and the goal.
Given the agent's coordinates, we consider a cone which spans around the beeline from the current position to the respective goal region. 
This implicitly assumes that for this environment the optimal solution is rather ``aligned'' with the beeline.
Clearly this particular choice works in this environment but would fail in others such as mazes. 
Despite one could construct pathological cases in which the ``factorizability'' of the problem is zero, we believe that a wide array of interesting applications allows to define useful factorization heuristics (e.g., reachability sets in autonomous driving, corridors in warehouses,...). 
In this scenario, we leveraged the assumption that agents will not have to reverse direction or leave the cone to find an optimal path goal.
\subsection{Results and Discussion}
The performance comparison between $\rrg$ and $\factrrg$ are presented in~\Cref{fig:costevolution}. In all the three instances one can appreciate that the \emph{factorized} algorithm outperforms the baseline by finding lower cost paths sooner, with a performance gap that grows as the number of agents increases.
Importantly, we remark that this happens in spite of a slight increase in computational cost per iteration. 
For instance, in the $4$-agent case we pass from $80 (\pm 11)\,ms$ to $98 (\pm13)\,ms$ ($+22\%$), but achieving a significant lower cost with many fewer iterations ($-75\%$).

We happened to observe that for low number of samples the standard algorithm could seldomly find a solution earlier than its factorized counterpart.
This happens due to the use of this factorization heuristic, which prevents $\factrrg$ to create connections from already factorized nodes to non-factorized ones (substantially exploiting the information provided by the heuristic: ``from here on the solutions live in the cone''). Since $\rrg$ does not have to respect the future resource constraint, it creates more edges of the graph for low number of samples, this is indeed reflected also in~\Cref{table:factgain} for the full graph.
Nevertheless, as the number of samples increases, the \emph{factorization} advantage becomes noticeable.

Lastly in~\Cref{table:factgain} we display the average dimension of the graphs at the end of each run for the two algorithms in the $3$-agent case. It emerges, that the \emph{factorized} algorithm $\factrrg$ is able to outperform $\rrg$ in terms of solution cost while also drastically reducing the graph dimension. 
The first reason for the edge number reduction is the fact that the graph created by $\rrg$ contains many bidirectional edges, nonetheless the edge number ratio is much smaller than $50 \%$. 
The second and most important reason is the following, suppose there exist $a$ single agent nodes each with $m$ outgoing edges, the number of edge required to express this graph if the agents were to be joint is $m^a$, exponential in the number of agents, while it is only linear ($am$) in the decoupled representation. 
\begin{table}[htb!]
\scriptsize
\centering
\begin{threeparttable}
    \captionsetup{labelsep=period, skip=5pt}
    \caption{}
    \label{table:factgain}
    \begin{tabular}{l@{\extracolsep{\fill}}*{4}{c}} 
     \toprule
      \textbf{\# agents} && \textbf{2}  & \textbf{3}  & \textbf{4} \\
      \midrule
     $\rrg$ ($\#$ edges) && 
     $90.4k(\pm2.8k)$ &
     $60.4k(\pm2.4k)$ &
     $34.6k(\pm5.9k)$ \\
     $\factrrg$ ($\#$ edges) &&
     $\mathbf{13.0k(\boldsymbol{\pm} 0.9k)}$ & 
     $\mathbf{9.7k(\boldsymbol{\pm} 0.7k)}$ &
     $\mathbf{8.3k(\boldsymbol{\pm} 0.8k)}$\\
     \bottomrule
     \end{tabular}
\end{threeparttable}
\end{table}
\section{Conclusion}
By applying the \emph{factorization} idea to standard~\acrshortpl{sba} we showed that one can naturally build factorized \acrshortpl{sba} that are much better suited for multi-agent planning problems.
Most importantly, the resulting algorithms often preserve the properties of the original algorithm (completeness, optimality, anytime,\ldots) while drastically reducing the computation load (shown analytically for $\prmstar$ and empirically for $\rrg$).

The effectiveness of the factorization depends on an heuristic function which is able to tell when players' solutions are decoupled in the remaining problem.
Even though we provided concrete examples, we identify this as one of the most interesting future research direction.
Another conceivable extension is showing that the factorization principle extends also to the more sophisticated \acrshortpl{sba}.
To conclude, we are intrigued that the full potential of factorization is yet to be discovered.
\bibliographystyle{./IEEEtrannourl}
\bibliography{IEEEabrv, references}
\section{Appendix}
\begin{definition}[$\ell_p$-dispersion]\label{def:dispersion}
For a finite, non-empty set $\pointset$ of points contained in a $d$-dimensional compact Euclidean
subspace $\statespace$ with positive Lebesgue measure, its $\ell_p$-dispersion $\disp_p(\pointset)$ is defined as
\begin{equation}
\disp_p(\pointset) = \sup_{\state \in \statespace}\left \{ \min_{s \in \pointset} \norm{x - p }_p \right \}.
\end{equation}
\end{definition}
Intuitively, the dispersion quantifies how well a space is covered by a set of points $\pointset$ in terms of the largest open Euclidean ball that touches none of the points.
\begin{proposition}[Sufficient Number of Samples for $\ell_\infty$-dispersion]\label{th:suffsamples}
Let $\statespace$ be a \emph{Lebesgue-measurable} space of measure $\lebesgue(\statespace)$ and dimension $d$. Let $\pointset$ be a sequence of $N$ i.i.d. uniform samples drawn on $\statespace$, for any $\dispbar, \pbar \in (0,1]$, we have that a sufficient number of samples $N>0$ such that $\prob{[\disp_\infty (\pointset) \leq \dispbar]} \geq \pbar$ is given by:
\begin{equation}
   N \geq \frac{\mu(\statespace)}{\dispbar^d} \log{\left( \frac{\mu(\statespace)}{\dispbar^d \left(1 - \pbar \right)}\right) }.
\end{equation}
\end{proposition}
\begin{proof}
    Subdivide $\statespace$ in $m=\ceil{\frac {\mu{(\statespace)}}{\dispbar^d}}$ identical $d$-dimensional hypercubes (cells). Note that the side of each cube is smaller or equal than $\dispbar$. 
    Moreover, note that having at least one sample per cell is a \emph{sufficient condition} for $\disp_\infty (\pointset) \leq \dispbar$.
    Let $\eventone_i$ be the event that the $i$-th cell does not contain any of the samples from $\pointset$, the probability of $\eventone_i$ is then $\prob{(\eventone_i)} = \left( 1 - \frac 1 m \right)^N$. Hence, by applying the union bound, the probability that at least one cell remains empty reads
    \begin{equation}
        \prob{[\cup_i \eventone_i]} \leq \sum_{i}\prob{[\eventone_i]} = m\left( 1 - \frac 1 m \right)^N.
    \end{equation}
    Therefore by considering the complementary event ``all the cells contain at least one sample''--which would ensure the desired discrepancy--we have:
    \begin{equation}
        \prob{[\disp_\infty (\pointset) \leq \dispbar]} \geq 1- m\left( 1 - \frac 1 m \right)^N \geq 1 - m \exp{(- \frac N m)},
    \end{equation}
    where the second inequality follows from $1 + x \leq e^x$ valid for any $x \in \reals$.
    Finally, for the given $\pbar$, substituting $m$, and solving for $N$ we obtain: 
    \begin{equation}
        N \geq  \frac{\mu(\statespace)}{\dispbar^d}\log{\left( \frac{\mu(\statespace)}{\dispbar^d \left(1 - \pbar \right)}\right),}
    \end{equation}
    where we use the fact that $1 - m \exp{(- \frac N m)}$ is decreasing in $m$ and $m=\ceil{\frac {\mu(\statespace)}{\dispbar^d}} \geq \frac{\mu(\statespace)}{\dispbar^d}$.
\end{proof}

\begin{lemma}[$\epsilon$-optimality of product of solutions] \label{thm:epsiloncomposition}
    Given a \acrshort{ma-mpp} let $\mypath^*$ be its optimal solution and consider $\mypath = \{\mypath_i \}_{i \in \agents}$ to be a feasible solution to the problem. Then:
    \begin{equation}
        \epsilon = \frac{\cost(\mypath) - \cost(\mypath^*)}{\cost(\mypath^*)} \leq \max_{i \in \agents}{(\epsilon^i)}
    \end{equation}
    where $\epsilon^i=\frac{\cost^i(\mypath) - \cost^i(\mypath^*)}{\cost^i(\mypath^*)}$.
\end{lemma}
\begin{proof}
    From the definition of joint cost and $\epsilon$-optimal: 
    \begin{equation}
    \begin{aligned}
        \epsilon &= \frac{\cost(\mypath) - \cost(\mypath^*)}{\cost(\mypath^*)} = \frac{\sum_{i \in \agents}\cost^i(\mypath) - \sum_{i \in \agents}\cost^i(\mypath^*)}{\sum_{i \in \agents}\cost^i(\mypath^*)} \\
        & = \frac{\sum_{i \in \agents} \left[ \cost^i(\mypath) - \cost^i(\mypath^*) \right]}{\sum_{i \in \agents}\cost^i(\mypath^*)} \leq \frac{\sum_{i \in \agents} \left[ \epsilon^i \cost^i(\mypath^*) \right]}{\sum_{i \in \agents}\cost^i(\mypath^*)} \leq \max_{i \in \agents}{\epsilon^i}.
    \end{aligned}
    \end{equation}
\end{proof}
\begin{lemma}[$\epsilon$-optimality of concatenation of paths] \label{thm:epsilonunion}
\Cref{thm:epsiloncomposition} applies also to concatenation of paths.
\end{lemma}
\begin{proof}
Analogous to~\Cref{thm:epsiloncomposition}.
\end{proof}

\end{document}